\documentclass{article}

\usepackage{microtype}
\usepackage{graphicx}
\usepackage{subfigure}
\usepackage{booktabs} 
\usepackage[utf8]{inputenc}
\usepackage{textgreek}
\usepackage{hyperref}

\usepackage[accepted]{icml2025}

\usepackage{amsmath}
\usepackage{amssymb}
\usepackage{mathtools}
\usepackage{amsthm}

\usepackage[capitalize,noabbrev]{cleveref}

\theoremstyle{plain}

\theoremstyle{definition}

\theoremstyle{remark}

\usepackage[textsize=tiny]{todonotes}

\icmltitlerunning{Humanoid World Models}

\begin{document}

\twocolumn[
\icmltitle{Humanoid World Models \includegraphics[height=1em]{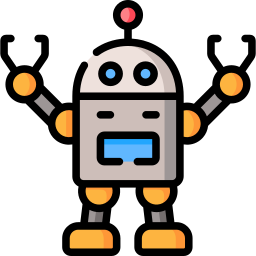} : Open World Foundation Models for Humanoid Robotics}



\icmlsetsymbol{equal}{*}

\begin{icmlauthorlist}
\icmlauthor{Qasim Ali}{equal,uw}
\icmlauthor{Aditya Sridhar}{equal,uw}
\icmlauthor{Shahbuland Matiana}{uw}
\icmlauthor{Alexander Wong}{uw}
\icmlauthor{Mohammad Al-Sharman}{uw}
\end{icmlauthorlist}

\icmlaffiliation{uw}{University of Waterloo, Waterloo, Canada}

\icmlcorrespondingauthor{Qasim Ali}{m45ali@uwaterloo.ca}
\icmlcorrespondingauthor{Aditya Sridhar}{a27sridh@uwaterloo.ca}

\icmlkeywords{Machine Learning, ICML}

\vskip 0.3in
]



\printAffiliationsAndNotice{\icmlEqualContribution} 

\begin{abstract}

Humanoid robots, with their human-like form, are uniquely suited for interacting in environments built for people. However, enabling humanoids to reason, plan, and act in complex open-world settings remains a challenge. World models, models that predict the future outcome of a given action, can support these capabilities by serving as a dynamics model in long-horizon planning and generating synthetic data for policy learning. We introduce Humanoid World Models (HWM), a family of lightweight, open-source models that forecast future egocentric video conditioned on humanoid control tokens. We train two types of generative models, Masked Transformers and Flow-Matching, on 100 hours of humanoid demonstrations. Additionally, we explore architectural variants with different attention mechanisms and parameter-sharing strategies. Our parameter-sharing techniques reduce model size by 33–53\% with minimal impact on performance or visual fidelity. HWMs are designed to be trained and deployed in practical academic and small-lab settings, such as 1–2 GPUs.

\end{abstract}

\section{Introduction}

Autonomous humanoid robots have the potential to transform both industry and daily life by automating tasks that are dull, dangerous, or physically demanding \cite{humanoid_ref}. Their human-like morphology allows them to operate in spaces built for people, interact naturally with humans, and easily learn from teleoperated demonstrations. However, in order to navigate the complexity and unpredictability of real-world environments these agents require sophisticated reasoning capabilities.

While large multimodal models \cite{ zhang2024visionlanguagemodelsvisiontasks, black2024pi0} show some promise in reasoning tasks, they often lack the accuracy, reliability, and robustness needed for embodied AI in open-world settings \cite{tong2024eyeswideshutexploring, embodiedai_challenges}. As a result, they struggle to meet the demands of embodied agents that must act safely and effectively in dynamic, unstructured environments \cite{embodiedai_challenges, li2024llmenhancedscenegraphlearning, duan2024ahavisionlanguagemodeldetectingreasoning}.

\begin{figure}[ht]
\centering
\centerline{\includegraphics[width=0.85\columnwidth]{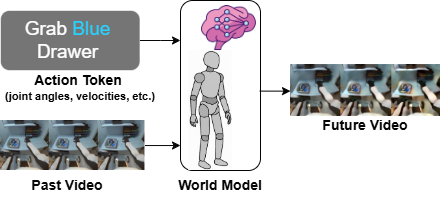}}
\caption{Overview of Humanoid World Models (HWM). Given past video observations and humanoid control tokens (joint angles, velocities, etc.), it predicts future video observations. }
\label{fig:overview}
\end{figure}

One avenue for improving humanoid intelligence and control is through \textit{World Models}—predictive models trained to forecast future outcomes based on past observations and actions \cite{haworldmodel}. In our setting, these function as action-conditioned video generators: they predict future visual states as sequences of frames, enabling agents to simulate the consequences of candidate actions \cite{du2023videolanguageplanning}. By simulating the outcomes of actions without trying them in the real world, world models can be used for long-horizon planning by using them as a dynamics model \cite{yang2024videonewlanguagerealworld} and enable more data-efficient policy learning through synthetic rollouts \cite{unisim}.

Despite recent progress in video generation, most models are built for entertainment applications \cite{liu2024sorareviewbackgroundtechnology, polyak2024moviegencastmedia}, emphasizing visual appeal over physical, ego-centric plausibility. Many world models remain closed-source, require large-scale compute for training or inference, or are not designed for humanoid robots \cite{unisim, nvidia2025cosmosworldfoundationmodel, FangqiIRASim2024}. As a result, there is a clear gap: few open-source models are both physically grounded for humanoid robots and lightweight enough to train and deploy on modest academic hardware (e.g., 2–3 GPUs). To address this, we ask: Can we build a physically plausible, humanoid-specific world model that can be trained and deployed on as little as two GPUs? 

We introduce \textbf{Humanoid World Models} (HWM), a set of open-source, lightweight world models for humanoid robotics trained on 100 hours of humanoid video demonstrations. We investigate two distinct video generation paradigms: Masked Transformers and Flow-Matching models. Drawing from recent advances in image and video generation, we explore four architectural variations within each framework. These variations are defined along two axes: (1) joint vs. cross-attention mechanisms, and (2) shared vs. separate parameters across token streams. This design space is motivated by successful architecture strategies from the more well-studied text-to-image model literature.

Our experiments show that, given our dataset and compute constraints, Masked Transformers consistently outperformed Flow-Matching models even when the latter used more parameters and were trained for longer. Across both families of models, we observed that the different architectural variants performed comparably in most scenarios. However, key trends emerged: within the Masked Transformer framework, the joint attention variant achieved the best overall performance. In contrast, for Flow-Matching models, split attention proved most effective. 

Importantly, we found that parameter-sharing strategies yielded near-identical performance to their non-shared counterparts while reducing parameter counts by 33–53\%, significantly lowering computational requirements. These results suggest that architectural and efficiency trade-offs—particularly attention design and parameter sharing—can be leveraged to build lightweight, performant world models without compromising quality.

\section*{Related Works}

\subsection{Humanoid Robots}

Humanoid robots (Humanoids) are actuated, bipedal, and bimanual robots designed to anthropomorphically resemble the human body structure \cite{humanoid_ref}. Their morphology is well-suited for operating in human-centered environments, enabling effective interaction and physical compatibility with everyday settings. This embodiment facilitates deployment across a wide range of real-world domains, including households \cite{humanoid_in_home}, manufacturing facilities \cite{humanoid_industrial}, elderly care homes \cite{humanoid_in_home}, and clinical or medical environments \cite{humanoid_ref}. Beyond physical compatibility with environments, the human-like form of humanoids allows for more natural interaction with people and the ability to imitate human behaviors \cite{humanoid_interaction}. Furthermore, collecting demonstration data is relatively straightforward for these platforms, as they can directly mimic human motions and tasks \cite{zhao2023learningfinegrainedbimanualmanipulation}. Our world models are trained specifically for humanoids, but can easily be extended to other embodiments.

\subsection{World Models}

A core challenge in robotics is enabling agents to perceive their environment, reason over possible actions, and plan goal-directed behavior, especially in novel or unstructured settings. Foundation models \cite{bommasani2022opportunitiesrisksfoundationmodels} attempt to generalize across tasks through large-scale, multimodal training. Recent approaches include Vision-Language-Action (VLA) models that predict actions from video and language inputs \cite{  black2024pi0, kim24openvla}, and large language or vision-language models used as high-level planners \cite{liang2023codepolicieslanguagemodel, li2024llmenhancedscenegraphlearning, wang2024llm3largelanguagemodelbasedtask}. However, these methods often struggle with spatial reasoning \cite{tong2024eyeswideshutexploring}, continuous sensorimotor processing, and require complex prompting strategies \cite{li2024llmenhancedscenegraphlearning}.

In contrast, video generation models offer a more grounded alternative for embodied agents. Video captures fine-grained physical and temporal structure that language alone cannot express \cite{yang2024videonewlanguagerealworld}. When trained to predict future video frames given past observations and actions, these models serve as \textit{World Models}—internal simulators that forecast the outcomes of potential action sequences \cite{haworldmodel, du2023videolanguageplanning}. This enables agents to plan through imagination, reason counterfactually, and generate synthetic experience for policy training.

In this work, we develop a humanoid-specific world model implemented as a video generator: it predicts future egocentric video frames from past video and action sequences. This generative modeling approach allows us to simulate plausible futures, supporting both planning and data-efficient learning in complex environments.

\subsection{Video Generation Models}

Early video generation relied on GANs \cite{goodfellow2014generativeadversarialnetworks}, but training instability limited their use. Recent methods improve both quality and efficiency by training generative models like diffusion models \cite{sohldickstein2015deepunsupervisedlearningusing, rombach2022highresolutionimagesynthesislatent} or masked transformers \cite{vaswani2023attentionneed} in the compressed latent spaces of Variational Autoencoders (VAEs) \cite{harvey2022conditionalimagegenerationconditioning}. Discrete latent approaches like vector quantized VAEs (VQ-VAEs) \cite{oord2018neuraldiscreterepresentationlearning} enable token-based video generation using masked or autoregressive transformers \cite{ramesh2021zeroshottexttoimagegeneration, chang2022maskgitmaskedgenerativeimage}. Flow Matching \cite{lipman2023flowmatchinggenerativemodeling} offers an alternative to diffusion with simpler training and faster sampling while preserving sample quality.

\textbf{Masked Video Generation}:  Transformer-based approaches have been widely used for image and video generation. These methods operate in the finite and quantized spaces of VQ-VAEs.  MaskGIT \cite{chang2022maskgitmaskedgenerativeimage} introduced a masked, bidirectional transformer for image generation from VQ-VAE latents, along with a non-autoregressive decoding scheme that significantly reduced sampling time compared to diffusion and autoregressive methods. Masked token prediction offers two key advantages over autoregressive approaches \cite{ramesh2021zeroshottexttoimagegeneration, wu2024ivideogpt}: (1) bidirectional context across space and time improves representation learning, and (2) tokens can be decoded in parallel, greatly accelerating inference. MAGVIT \cite{yu2023magvitmaskedgenerativevideo} extended this approach to video using spatio-temporal VQ-VAEs, while MAGVIT2 \cite{yu2024languagemodelbeatsdiffusion} introduced a stronger tokenizer that outperformed diffusion-based baselines with fewer sampling steps. Open-MAGVIT2 \cite{luo2024openmagvit2opensourceprojectdemocratizing} provides an open-source implementation. We explore similar non-autoregressive masked video transformers for building humanoid-specific world models.

\textbf{Diffusion and Flow-Matching}: Video diffusion models \cite{ho2022videodiffusionmodels} extend diffusion processes to sequences of images, modeling both spatial and temporal dynamics. However, the added temporal dimension significantly increases computational cost. Recent advances in spatiotemporal VAEs \cite{surveyvideodiffusionmodels, bartal2024lumierespacetimediffusionmodel} mitigate this by compressing video into low-dimensional latent spaces, making training and inference more tractable.

Large-scale text-to-video diffusion models such as Sora \cite{liu2024sorareviewbackgroundtechnology}, CogVideoX \cite{yang2025cogvideoxtexttovideodiffusionmodels}, and MovieGen \cite{polyak2024moviegencastmedia} achieve impressive visual quality but are designed for entertainment and lack support for conditioning on past video, a key requirement for physically grounded, ego-centric prediction.

Several recent works explore video generation for robotic agents, but most are not designed for humanoid platforms and are not open source. UniSim \cite{unisim} trains a text-conditioned video diffusion model for zero-shot policy transfer, but it relies on an outdated U-Net backbone and is not open source. IraSim \cite{FangqiIRASim2024} uses a factorized spatial-temporal transformer within a diffusion framework, but targets robot arms and lacks temporal compression, relying instead on frame-wise image VAEs. Navigation World Models \cite{bar2025navigationworldmodels} are limited to low-DoF mobile robots and generate individual future frames rather than continuous video sequences.

NVIDIA’s Cosmos \cite{nvidia2025cosmosworldfoundationmodel} is a notable exception—an open-source, high-fidelity video-to-video model. However, it is not designed for humanoid embodiments and is prohibitively resource-intensive. Its smallest variant (7B parameters) requires 8 NVIDIA H100 GPUs for training and over 40GB of VRAM for inference. On our compute setup of 2 NVIDIA A6000s, generating 121 frames video took over an hour, making finetuning or deployment impractical.

We explore training a lightweight flow-matching model in the continuous latent space of Cosmos’s VAE \cite{nvidia2025cosmosworldfoundationmodel}.

\subsection{Video Transformer Architectures}

For the Masked Video Model, we follow prior work in using factorized spatio-temporal attention \cite{bruce2024genie, xiang2024pandorageneralworldmodel} to reduce computational overhead. This approach separates spatial and temporal attention to scale more efficiently with video length.

In the broader diffusion and flow-matching video generation models literature, architectural trends have shifted from convolutional U-Nets \cite{ho2020denoisingdiffusionprobabilisticmodels} to transformer-based designs for improved scalability \cite{peebles2023scalablediffusionmodelstransformers}. In image generation, joint attention blocks—where image and context tokens are processed together—are now standard, as seen in Stable Diffusion 3 (SD3) \cite{esser2024scalingrectifiedflowtransformers}. Subsequent work has shown that complexity can be further reduced through parameter sharing across token streams \cite{fal_auraflow, chen2025ditairrevisitingefficiencydiffusion}.

In contrast, most video diffusion models still avoid joint attention due to its high memory cost over long video sequences. Instead, they typically use a two-stage attention scheme: self-attention over video tokens, followed by cross-attention with context \cite{polyak2024moviegencastmedia, nvidia2025cosmosworldfoundationmodel}. Leading models such as Cosmos \cite{nvidia2025cosmosworldfoundationmodel} follow this structure, while more complex designs like mixture-of-experts \cite{kong2025hunyuanvideosystematicframeworklarge} or pyramidal transformers \cite{jin2024pyramidalflowmatchingefficient} further increase system complexity.

In our work, we revisit joint attention for video generation. Joint attention over spatiotemporal video tokens is now feasible thanks to two key advances: (1) highly compressive VAEs that reduce spatial resolution (e.g. factor of 16) and temporal resolution (e.g. factor of 8), and (2) efficient parameter-sharing techniques from recent image generation models \cite{fal_auraflow, chen2025ditairrevisitingefficiencydiffusion}. 

\section{Methodology}

We develop two humanoid-specific world models based on distinct generative paradigms: \textit{Masked Humanoid World Model} (Masked-HWM) and \textit{Flow Humanoid World Model} (Flow-HWM). Masked-HWM employs masked video modeling in a discrete latent space (via VQ-VAE), while Flow-HWM uses flow matching in a continuous latent space. We detail the video generation frameworks and architectures in \ref{sec:hwm-mvm} and \ref{sec:Flow-HWM} respectively, but describe the transformer block design in detail in \ref{sec:block-design}.

\subsection{Formulation}
The goal is to predict plausible future video frames given a sequence of past video frames and associated actions. Formally, the model predicts a sequence of $f$ future RGB frames $v_f \in \mathbb{R}^{f \times 3 \times H \times W}$, conditioned on $p$ past frames $v_p \in \mathbb{R}^{p \times 3 \times H \times W}$, $p$ past actions $a_p \in \mathbb{R}^z$, and $f$ future actions $a_f \in \mathbb{R}^z$. Action vectors include joint angles, velocities, and gripper states of the humanoid.

Following prior works \cite{rombach2022highresolutionimagesynthesislatent, chang2022maskgit}, we train our generative models in a VAE's compressed latent space: $v_p$ and $v_f$ are encoded into latent representations $L_p$ and $L_f$. Masked-HWM uses a VQ-VAE to quantize latents into tokens from a finite vocabulary of size $s$, while Flow-HWM uses a continuous VAE.

\subsection{Masked Video Modelling} \label{sec:hwm-mvm}
We train the Masked-HWM variant using the Masked Video Modelling (MVM) paradigm, inspired by MaskGIT and MAGVIT \cite{chang2022maskgit, yu2023magvitmaskedgenerativevideo}. After passing passing $v_p, v_f$ through a VQ-VAE to yield \( \mathbf{L_p}, \mathbf{L_f} \), we concatenate the past and future latent tokens \( \mathbf{L} = [\mathbf{L_p}; \mathbf{L_f}]\) along the temporal dimension. 

During training, Masked-HWM receives corrupted and masked versions of the latent sequence as input. Following Copilot-4D \cite{zhang2023copilot4d}, we add noise to the latents by corrupting \(\mathbf{L}\) with random token replacements at a rate uniformly sampled from \(\mathcal{U}(0, \rho_{\text{max}})\), where \( \rho_{\text{max}} \) denotes the maximum corruption rate. Next, we apply masking to the future latents $\mathbf{L_f}$ using a per-frame thresholding strategy. For each frame, we sample a value $r \sim \mathcal{U}(0, 1)$ and compute a masking threshold $\gamma(r)$ using a predefined scheduling function. Then, for each token in the future sequence $\mathbf{L_f}$, we sample a probability from $\mathcal{U}(0, 1)$ and mask the token if it falls below the frame's threshold. 

The model is trained to reconstruct the original tokens at these masked positions. Let $\mathbf{M}$ denote the binary mask indicating which tokens have been masked within corrupted $L_f$. The training objective is to minimize the cross-entropy loss between the predicted tokens \( \hat{\mathbf{L}}_f \) at the masked locations \( \mathbf{M} \) and the true tokens at those same locations, as follows:

\[
\mathcal{L} = - \mathbb{E}_{\mathbf{M}} \left[ \sum_{i} \mathbf{M}_{i} \log p(\hat{L}_i \mid \mathbf{L_f}) \right]
\]

where $\hat{L}_i$ is the model’s prediction and $\mathbf{M}_i \in {0, 1}$ indicates whether location $i$ was masked. This loss encourages the model to accurately reconstruct corrupted or hidden tokens in the future sequence.

At inference time, we begin by masking all tokens in the future latent sequence $\mathbf{L_f}$. Generation proceeds latent frame by latent frame: after predicting one frame's tokens, the result is fed back to help condition the next. Within each frame, tokens are predicted in parallel over $K$ refinement steps. At each step, a random subset of tokens is re-masked and re-predicted, allowing the model to iteratively improve its guesses. This parallel decoding strategy significantly accelerates generation compared to traditional autoregressive methods, which decode tokens sequentially. For more details on the sampling procedure, we refer readers to MaskGiT \cite{chang2022maskgit}. 

\subsubsection{Architecture}

\textbf{Tokenization} The input video frames are encoded using a VQ-VAE performing spatiotemporal compression, yielding discrete tokens. Each latent pixel is treated as a token and projected into $h$-dimensions. Action sequences ($a_p$, $a_f$) are independently embedded into the same $h$-dimensional space using a multi-layer perceptron (MLP), ensuring compatibility with the video tokens. \\
\textbf{Transformer}: The resulting four token streams—past video ($v_p$), future video ($v_f$), past actions ($a_p$), and future actions ($a_f$)—are fed into a stack of $d$ transformer blocks. After processing, the video tokens are linearly projected to a distribution over the VQ-VAE codebook of size $s$, representing the predicted token identities. Details of the transformer block variants are provided in Section~\ref{sec:block-design}. 

\subsection{Flow Matching for Video Generation} \label{sec:Flow-HWM}
\label{sec:flow-matching}

We train the Flow-HWM variant using the Flow Matching (FM) framework \cite{lipman2023flowmatchinggenerativemodeling, albergo2023stochasticinterpolantsunifyingframework, liu2022flowstraightfastlearning}. The framework formulates video generation as a continuous transformation of samples from a simple prior distribution (Gaussian noise) into data samples drawn from the target distribution. As opposed to learning a reversed stochastic process like in traditional diffusion models \cite{song2021scorebasedgenerativemodelingstochastic}, FM directly learns a time-dependent velocity field that drives this transformation.

Let $\mathbf{X}_1$ denote a video sample in the latent space, and let $\mathbf{X}_0 \sim \mathcal{N}(0, \mathbf{I})$ represent a random sample from the Gaussian prior. We train the model by sampling an intermediate time $t \in [0, 1]$ and construct a point along the trajectory $\mathbf{X}_t$ using linear interpolation:

\begin{equation}
    \mathbf{X}_t = t \mathbf{X}_1 + (1 - (1 - \sigma_{\text{min}}) t)\mathbf{X}_0,
\end{equation}

where $\sigma_{\text{min}}$ is a small positive constant ensuring non-zero support at $t = 1$. The ground-truth velocity of the transformation path is then given by the time derivative:

\begin{equation}
    \mathbf{V}_t = \frac{d \mathbf{X}_t}{dt} = \mathbf{X}_1 - (1 - \sigma_{\text{min}}) \mathbf{X}_0.
\end{equation}

Our model, parameterized by $\theta$, predicts the instantaneous velocity field $u_\theta(\mathbf{X}_t, \mathbf{P}, t)$ conditioned on the  past video frames $v_p$, past actions $a_p$, future actions $a_f$ $\mathbf{P}$ and time $t$. The training objective of the model is to minimize the expected mean squared error between the predicted and ground-truth velocity:

\begin{equation}
    \mathbb{E}_{t, \mathbf{X}_0, \mathbf{X}_1, a_p, a_f, v_p}= \left[ \left\| u_\theta(\mathbf{X}_t, a_p, a_f, v_p, t) - \mathbf{V}_t \right\|^2 \right].
\end{equation}

We adopt classifier-free guidance \cite{ho2022classifierfreediffusionguidance} to improve conditional generation by enabling the model to better balance conditioning signals from actions and past context during training and inference.
During inference, generation proceeds by integrating the learned velocity field from $t=0$ to $t=1$, starting from pure Gaussian noise and employing the first-order Euler ODE solver.

\subsubsection{Architecture}

\textbf{Tokenization} We tokenize the compressed latent video frames $L_p$ and $L_f$ by dividing them into $p_{lw} \times p_{lw}$ spatial and $p_{t}$ temporal segments per token. Each token is projected to $h$ channels via a convolutional layer. Action sequences $a_p$ and $a_f$ are embedded using an MLP into the same $h$-dimensional space. The timestep $t$ is encoded using sinusoidal embeddings following DDPM \cite{ho2020denoisingdiffusionprobabilisticmodels}.

\textbf{Transformer} After tokenization, each of the four streams of tokens ($v_f$,$v_p$,$a_f$,$a_p$) are kept separate and processed by $d$ transformer blocks sequentially.  In the final layer, we apply time modulation as described in DDPM, followed by a linear projection of the future tokens $v_f$ from $h$ dimensions back to $l$ latent dimensions. The resulting tokens are then reshaped into the original video's spatiotemporal format to be decoded back to pixel space the VAE. We detail the transformer block design in \ref{sec:block-design}.

\subsection{Transformer Block Design} \label{sec:block-design}
We evaluate several architectural variants of the transformer block used in both Masked-HWM and Flow-HWM, focusing on three key design dimensions: (1) attention structure (joint vs. split attention), (2) parameter sharing across token streams, and (3) token stream grouping (modality-based vs. fully separate).
\begin{figure}[ht]
\centering
\centerline{\includegraphics[width=0.8\columnwidth]{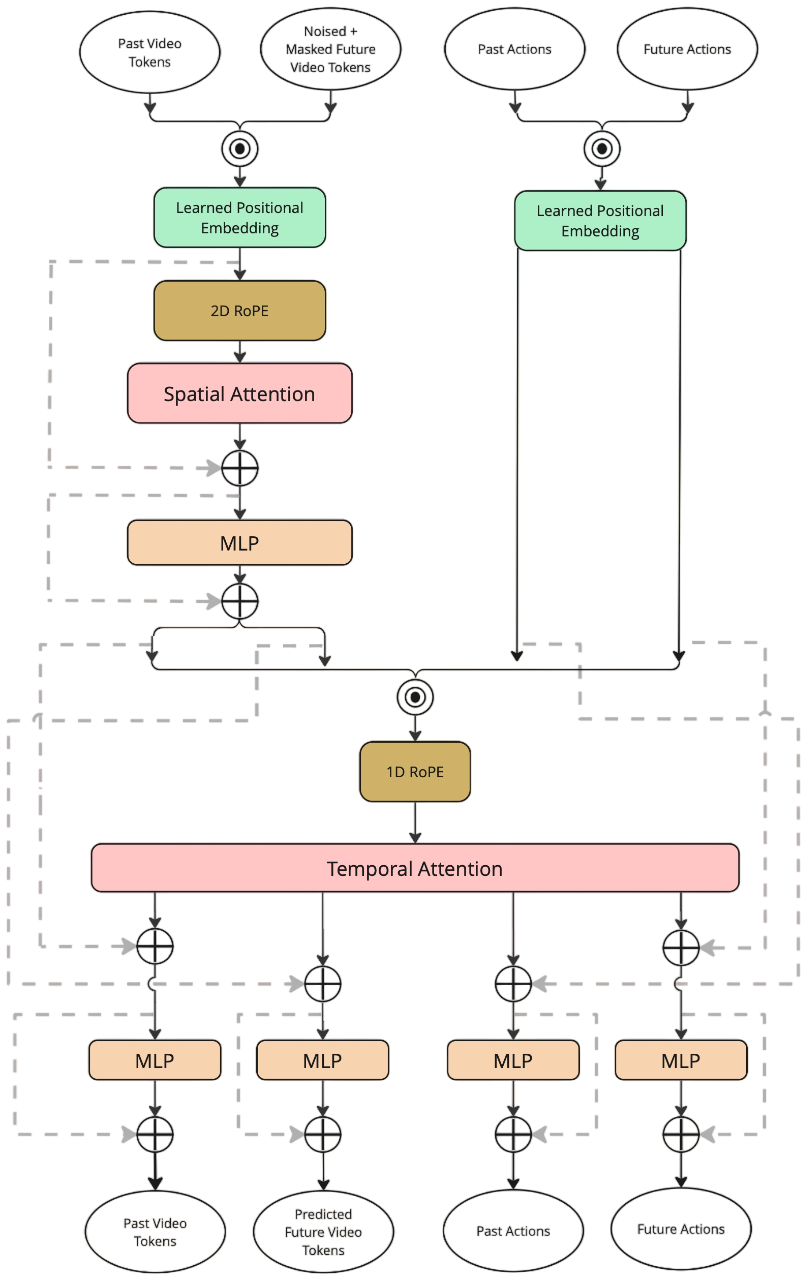}}
\caption{Architecture of a single transformer block in Masked-HWM (Base Block variant). Video and action streams are processed independently, with video streams also receiving Spatial Attention. All streams interact via joint Temporal Attention. RoPE is applied per attention type (2D for spatial and 1D for temporal). Each stream uses distinct MLP weights in the feedforward stage.} 
\label{fig:mvm_architecture}
\end{figure}

\begin{figure}[ht]
\centering
\centerline{\includegraphics[width=0.8\columnwidth]{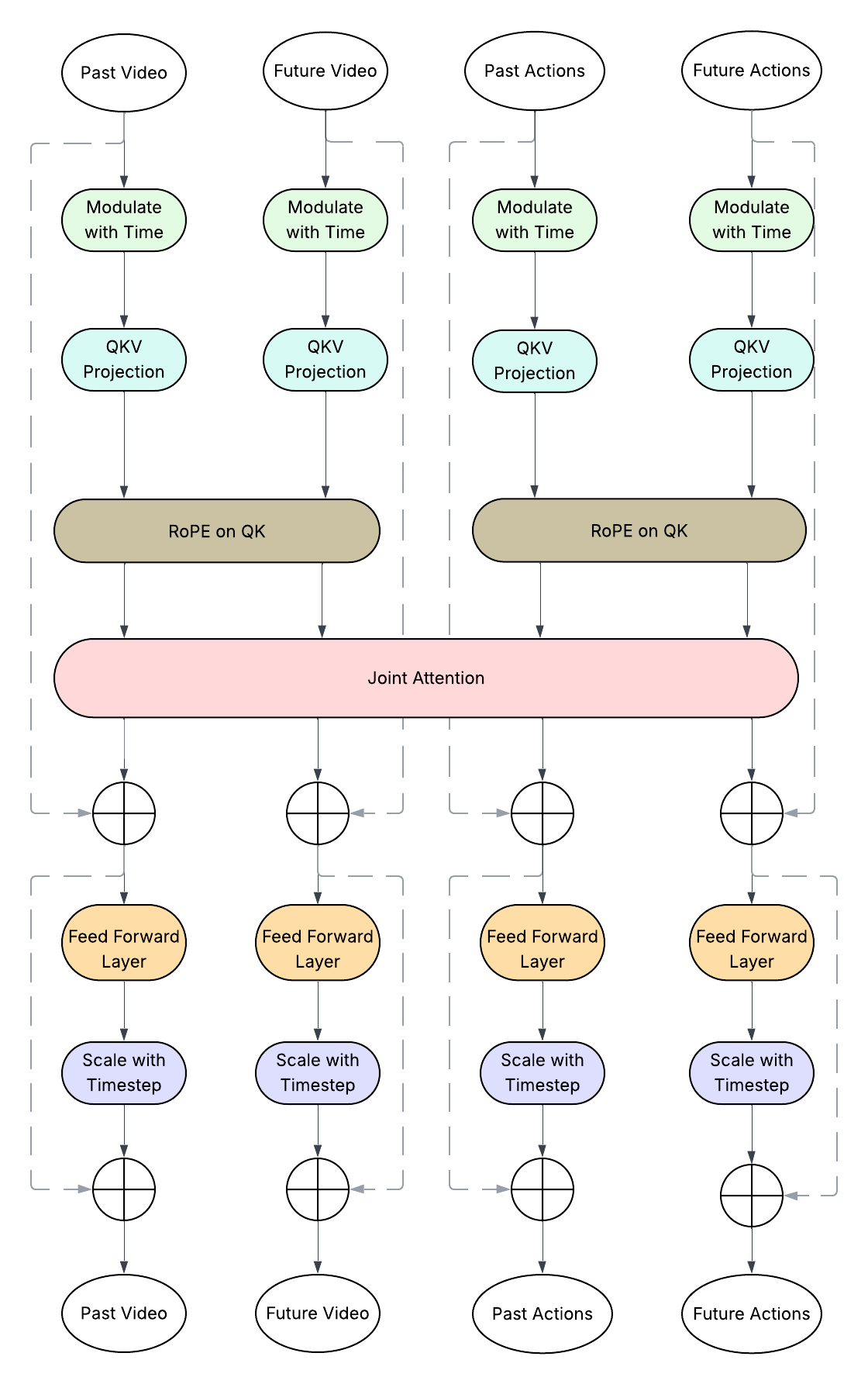}}
\caption{Architecture of a single transformer block in Flow-HWM (Base Block variant). Each token stream (past/future video and actions) uses separate weights for timestep modulation, QKV projection, and feedforward MLPs. Joint Attention integrates all streams. RoPE is applied by modality: 3D for video tokens, 1D for action tokens.}
\label{fig:flow_transformer}
\end{figure}

\textbf{Base Block}: We start by describing the Base Transformer Block, which is augmented to create other block designs. It processes four token streams—past video ($v_p$), future video ($v_f$), past actions ($a_p$), and future actions ($a_f$)—each with its own set of parameters but a shared joint attention layer.

The Base Block design for the Masked-HWM is illustrated in Figure~\ref{fig:mvm_architecture}. Following prior work in non-autoregressive video generation methods \cite{bruce2024genie, xiang2024pandorageneralworldmodel}, we adopt a factorized or separate spatial and temporal attention layers. Compared to full spatiotemporal attention, factorized attention reduces the computational cost and scales more efficiently with video length. During temporal attention, all tokens from various streams \([a_p, a_f, \mathbf{L}]\) jointly attend to one another along the temporal dimension. Spatial attention is applied seperately only to the video tokens. Rotary Position Embeddings (RoPE)\cite{su2023roformerenhancedtransformerrotary} are used during spatial and temporal attention.

The Base Block for the Flow-HWM variant, as illustrated in Figure~\ref{fig:flow_transformer}, is inspired by Stable Diffusion 3 (SD3)\cite{esser2024scalingrectifiedflowtransformers}. This design processes the four token streams—$v_p$, $v_f$, $a_p$, and $a_f$—with separate parameters and enables interaction through a joint attention operation. 
Within each block, each stream is first modulated by the timestep using learned scale $\alpha_0$ and shift $\beta_0$ parameters \cite{peebles2023scalablediffusionmodelstransformers}. Subsequently, stream-specific queries, keys, and values are computed using separate $W_{QKV}$ projections. We add positional encodings to the queries and keys of each token steam. Specifically, we add 3D Rotary Position Embeddings (RoPE) to the video tokens, as done in Cosmos \cite{nvidia2025cosmosworldfoundationmodel}, and apply 1D RoPE across time for the action tokens. Past and future tokens are concatenated prior to adding positional encoding.

A joint multi-stream attention operation is then applied across all tokens, followed by another timestep-dependent rescaling using $\gamma_0$. In the feedforward stage, tokens are again modulated with the timestep embedding using new parameters $\alpha_1$ and $\beta_1$, passed through a stream-specific MLP, and finally rescaled using $\gamma_1$. Residual connections are added during the attention and the feedforward stages.

\textbf{Parameter Sharing}: Recent advances in efficient diffusion transformer design \cite{chen2025ditairrevisitingefficiencydiffusion, fal_auraflow} have showcased that the benefits of joint attention can be garnered with far fewer parameters using shared attention. We evaluate two parameter-sharing strategies. These strategies selectively share key transformer components across token streams, including timestep modulation parameters ($\alpha$, $\beta$, $\gamma$), QKV projection weights ($W_{QKV}$), and feedforward MLPs.

In the \textit{Full Sharing} variant, the modulation scalars, QKV projections, and MLPs are shared across all four token streams ($v_p$, $v_f$, $a_p$, $a_f$). This maximally reduces parameter count and compute overhead. In the \textit{Modality Sharing} variant, parameters are shared within each modality, i.e., video streams ($v_p$, $v_f$) share weights, and action streams ($a_p$, $a_f$) share weights. This strikes a balance between model compactness and representational flexibility.

In our implementation, we retain the original \textit{Base Block} configuration (with fully separate parameters) for the first four layers and apply parameter sharing in the remaining $l - 4$ transformer layers, following practices from \cite{fal_auraflow}. This hybrid scheme allows early layers to learn modality-specific representations, while deeper layers focus on cross-modal reasoning in a more compact parameter regime.

\textbf{Split Attention.} While joint attention across all token streams enables rich cross-modal interactions, it becomes increasingly expensive with longer video sequences and higher token counts. To address this, we implement a two-stage \textit{Split Attention} mechanism, which has been widely adopted in recent large-scale video generation models \cite{nvidia2025cosmosworldfoundationmodel, bar2025navigationworldmodels, polyak2024moviegencastmedia} for its computational efficiency.

In this variant, each stream—future video ($v_f$), past video ($v_p$), past actions ($a_p$), and future actions ($a_f$)—first undergoes \textit{independent self-attention} within its own sequence. This allows each modality and temporal context to process intra-stream dependencies with minimal overhead.

Following self-attention, we apply a \textit{cross-attention} operation in which the future video tokens $v_f$ serve as queries, and the keys and values are drawn from the remaining streams ($v_p$, $a_p$, $a_f$). This structure enables the model to selectively condition future video generation on past observations and intended actions, while avoiding the full cost of global attention. As in the joint attention variant, both self- and cross-attention layers are modulated using time-dependent scaling and shifting, with parameters ($\alpha$, $\beta$, $\gamma$) learned per stream.

\section{Results}

\begin{table*}[t]
\caption{Performance of Masked-HWM variants. Base Block yields the best FID; Split Attention gives the highest PSNR. Parameter sharing improves efficiency with minimal quality loss. }
\label{tab:mvm_ablation_results}
\vskip 0.15in
\begin{center}
\begin{small}
\begin{sc}
\begin{tabular}{lcccc}
\toprule
Metric & Split Attention & Base Block & Modality Sharing & Full Sharing \\
\midrule
Model Size (Billion)     & 0.220 & 0.321 & 0.237 & \textbf{0.195} \\
Peak GPU Memory (GB)     & 2.22  & 2.63  & 2.30  & \textbf{2.12}  \\
Samples per second       & 2.09  & 2.27  & 2.25  & \textbf{2.36}  \\
FID                      & 15.31 & \textbf{10.13} & 11.67 & 14.21 \\
PSNR (dB)                & \textbf{29.37} & 29.02 & 28.97 & 28.66 \\
\bottomrule
\end{tabular}
\end{sc}
\end{small}
\end{center}
\vskip -0.1in
\end{table*}

\textbf{Dataset} We train our models on the 1xGPT dataset \cite{1X_Technologies_1X_World_Model_2024}, which contains 100 hours of egocentric video captured from the Humanoid EVE Android executing various tasks. Video frames are recorded at 30 Hz, with each frame paired with a corresponding action vector $a \in \mathbb{R}^{25}$ representing movement velocities, hand closure states, and pitch-yaw-roll angles of the joints (wrist, knee, elbow, shoulder, neck, and hip). We train both models to generate $f = 8$ future frames conditioned on $p = 9$ past frames at $H=256 \times W=256$ resolution.

\textbf{Evaluation} We evaluate our models using Fréchet Inception Distance (FID) \cite{fid} and PSNR \cite{psnr}, computed over 21,000 generated frames from a held-out validation set. To isolate the impact of the generative model from the VAE's reconstruction quality, all ground truth frames are passed through the same VAE encoder-decoder used during training. In addition to image quality, we report model parameter count, video latent decoding speed (measured in samples per second), and peak GPU memory usage during inference.

\subsection{Masked Video Modelling}
\textbf{Experimental Setup} We tokenize all video frames using the NVIDIA Cosmos DV8x8x8 tokenizer, which applies $8\times$ compression in both spatial and temporal dimensions, reducing $256 \times 256$ RGB frames to $32 \times 32$ latent grids. Each training sample consists of 2 fully unmasked past latents and 1 partially masked future latent. We use a cosine masking schedule from MaskGIT and inject Copilot-4D style noise with a uniform corruption rate sampled from $\mathcal{U}(0, \rho_{\text{max}})$, where $\rho_{\text{max}}=0.2$.

Models are trained for 60{,}000 steps using the AdamW optimizer, with a learning rate linearly decayed from $3\mathrm{e}{-5}$ after 100 warmup steps. We use 24 transformer layers, 8 heads, 512-dimensional tokens, and an MLP hidden size of 2048. We apply standard normal initialization ($\mu=0$, $\sigma=0.02$) for all weights, except Xavier initialization for the mask token and output projection. Training is performed on a single NVIDIA A6000 with batch size 16. During inference, we use $K=2$ decoding iterations.

\subsubsection{Qualitative Results} \label{sec:mvm_qualitative}
\begin{figure}[ht]
\centering
\centerline{\includegraphics[width=0.8\columnwidth]{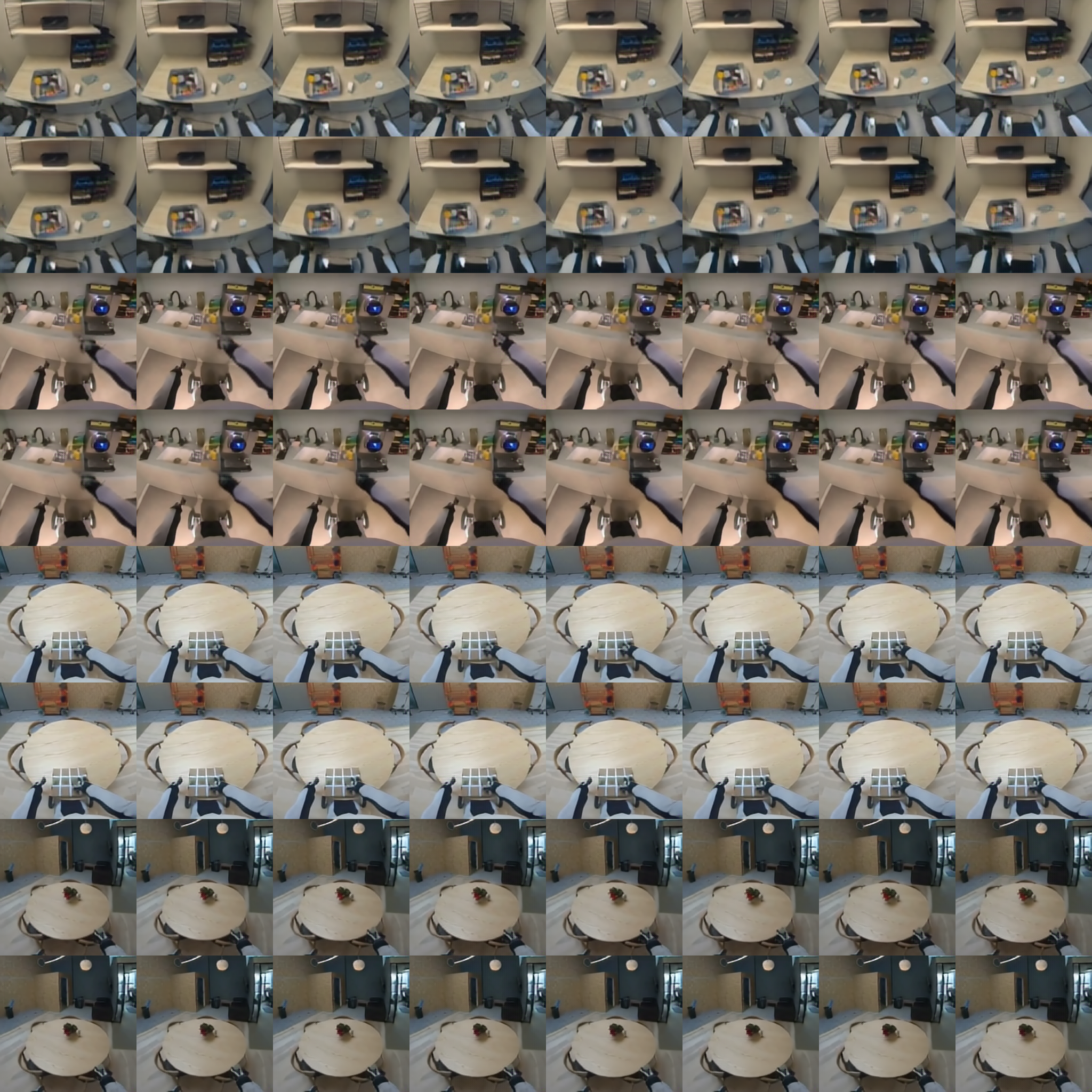}}
\caption{Four sample videos from the Base Block Variant of Masked-HWM.  Top row: generated frames; bottom row: ground truth.}
\label{fig:mvm_sample}
\end{figure}

Sample videos from the Base Block variant are shown in Figure \ref{fig:mvm_sample}. The model learns both structural elements of the scene, such as furniture and small objects that the robot is manipulating, and overall textures. Larger parts of the robot's appendage, including arms and wheels, are generally modeled accurately. However, precision elements, such as fingers, are often slightly blurry and entangled.  Further, the visual quality of generated images is robust to lighting, as shown in the 3rd (lighter) and 4th (darker) sequences.

\subsubsection{Quantitative Results}\label{sec:mvm_quantitative}
Table~\ref{tab:mvm_ablation_results} reports the performance of different Masked-HWM variants. The \textit{Base Block} achieved the best FID score (10.13), indicating the strongest visual quality among all configurations. The \textit{Split Attention} variant obtained the highest PSNR (29.37 dB), but slightly underperformed in FID, suggesting better pixel-level fidelity at the cost of less realistic global structure.

Both parameter-sharing variants—\textit{Modality Sharing} and \textit{Full Sharing}—reduced model size and memory while maintaining competitive FID and PSNR. \textit{Modality Sharing} matched Base Block quality with 26\% fewer parameters, showing that intra-modality sharing suffices. \textit{Full Sharing}, though slightly lower in quality, had the smallest footprint and fastest inference (2.36 samples/sec), making it ideal for efficiency-focused settings.

\begin{table*}[t]
\caption{Performance of Flow-HWM variants. Full Sharing achieves the best FID, memory usage, and speed even when compared to the Base Variant; Split Attention yields the highest PSNR.}
\label{tab:flow_model_perf_accuracy}
\vskip 0.15in
\begin{center}
\begin{small}
\begin{sc}
\begin{tabular}{lcccc}
\toprule
Metric & Split Attention & Base Block & Modality Sharing & Full Sharing \\
\midrule
Model Size (Billion)       & 0.944 & 1.36  & 0.886 & \textbf{0.648} \\
Peak GPU Memory (GB)       & 4.37  & 5.94  & 4.41  & \textbf{3.25}  \\
Samples per second         & 1.11  & 1.69  & 1.89  & \textbf{1.91}  \\
FID                        & 111.12  & 111.59   & 112.75  & \textbf{110.73}  \\
PSNR (dB)                  & \textbf{20.50}    & 20.42    & \textbf{20.50}   & 20.43    \\
\bottomrule
\end{tabular}
\end{sc}
\end{small}
\end{center}
\vskip -0.1in
\end{table*}

\subsection{Flow Matching}

\textbf{Experimental Setup.}

We use the Cosmos Continuous 8x16x16 tokenizer that spatially compresses frames by a factor of 16 (from $ 256\times256$ to $ 16\times16$), and performs $8\times$ temporal compression. Using a more spatially compressive VAE relative to Masked-HWM allows for joint attention while using much larger models required for flow-matching networks. Models are trained with $d=17$ transformer layers and $h=1172$-dimensional tokens using the AdamW optimizer, a learning rate of $1\mathrm{e}{-4}$, cosine learning rate scheduler, and batch size 128. Training runs for 150{,}000 steps across 2 NVIDIA A6000 GPUs. We use patch sizes $p_{lw}=2, p_t=1$. We initialize final linear layers with Xavier initialization, as zero-initialization  (as in prior works \cite{peebles2023scalablediffusionmodelstransformers}) led to instability. No learning rate warmup is used, as it degraded convergence. During inference, we apply 50 denoising steps with a classifier-free guidance scale of 3.0.

\subsubsection{Qualitative Results}

\begin{figure}[ht]
    \centering
    \includegraphics[width=0.8\columnwidth]{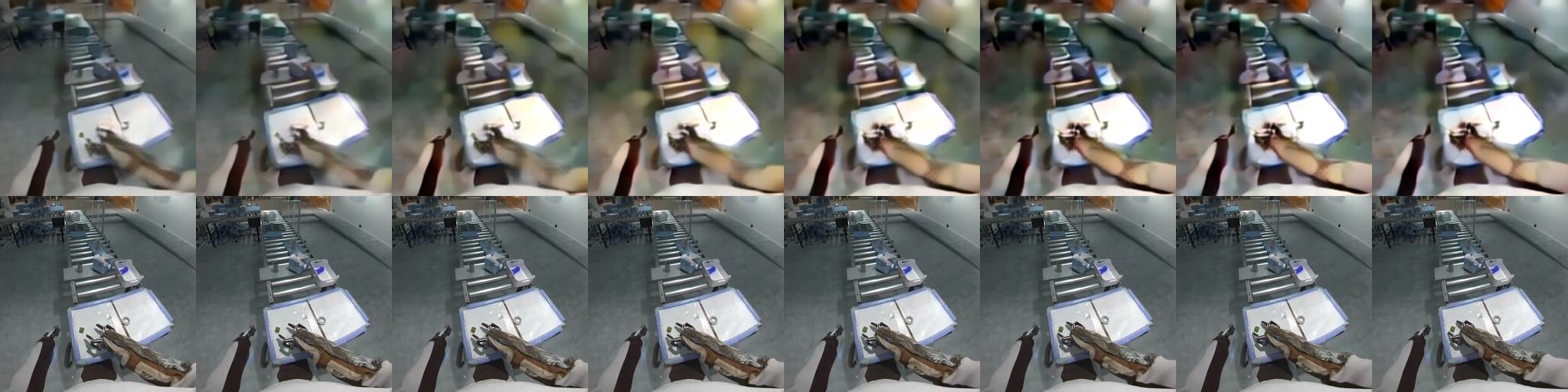}
    \includegraphics[width=0.8\columnwidth]{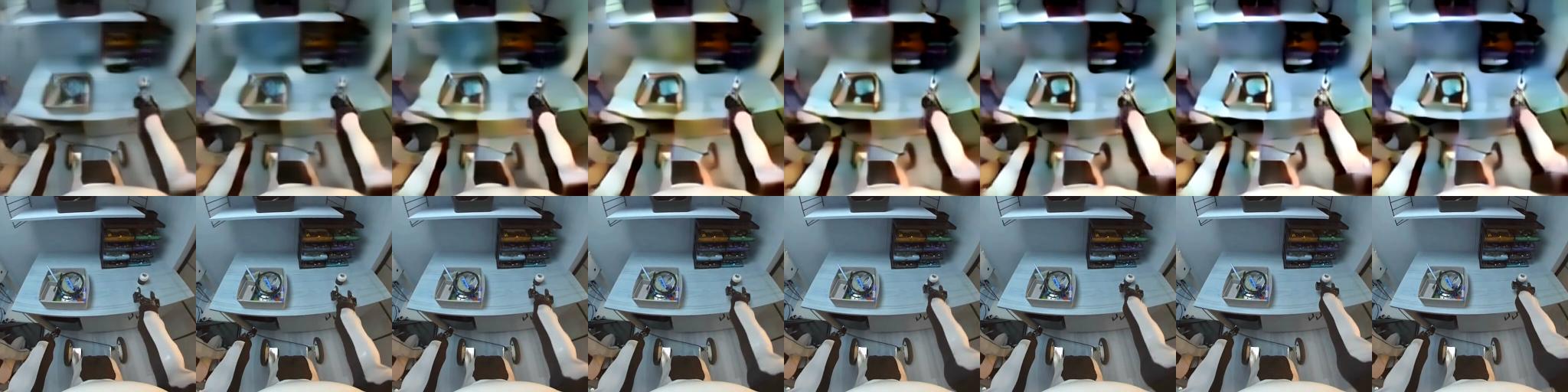}
    \includegraphics[width=0.8\columnwidth]{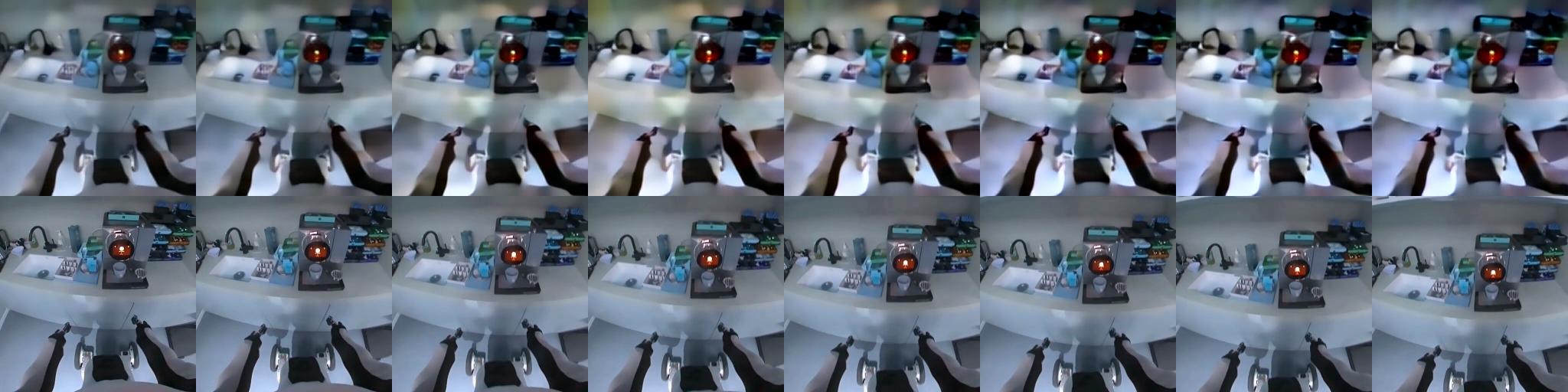}
    \includegraphics[width=0.8\columnwidth]{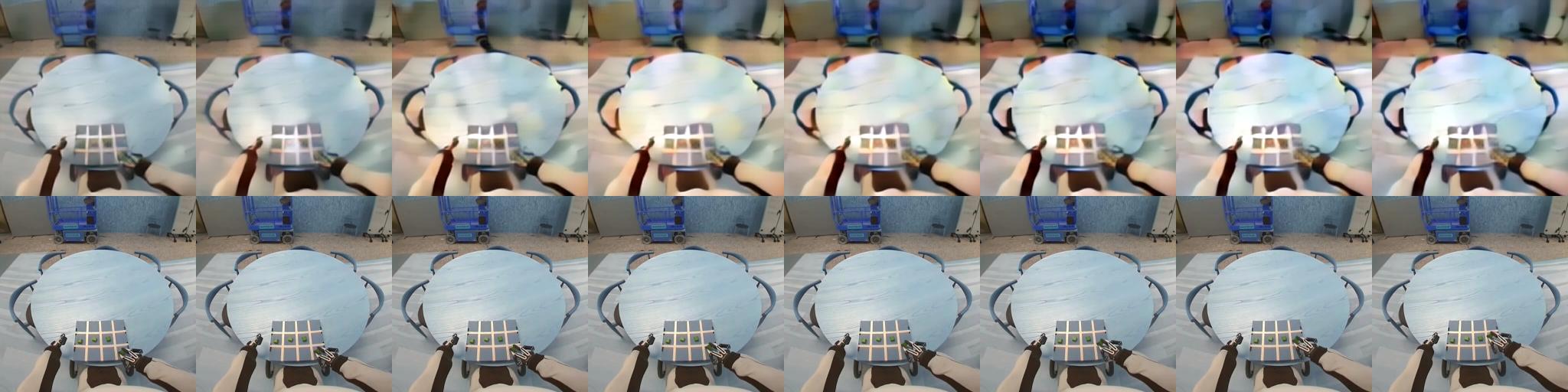}
    \caption{Sample videos from the Base Variant of Flow-HWM. Top row: generated frames; bottom row: ground truth.}
    \label{fig:flow_samples}
\end{figure}

As shown in Figure~\ref{fig:flow_samples}, the generated videos successfully capture overall scene structure (walls, floors, and doors). However, the outputs exhibit noticeable blurriness and artifacting like spotted patches. Visual quality tends to degrade in later frames, with the model struggling to preserve straight edges and rounded shapes. While the arms are often rendered with high fidelity, the model defaults to a canonical arm appearance and fails to represent unusual or out-of-distribution arm configurations (as seen in the top video strip).

\subsubsection{Quantitative Results}

Table~\ref{tab:flow_model_perf_accuracy} summarizes Flow-HWM performance. \textit{Full Sharing} offered the best trade-off, outperforming the Base Block in all metrics while halving the parameter count. It achieved the lowest FID (110.73), fastest inference (1.91 samples/sec), and minimal memory use (3.25 GB). \textit{Modality Sharing} delivered similar quality with notable efficiency gains, while \textit{Split Attention} yielded the highest PSNR (20.50 dB) but required more memory and was slower. Overall, parameter sharing improves efficiency without compromising quality.

Overall, none of the Flow-HWM variants outperformed the Masked-HWM models in either visual quality or sampling speed, suggesting that masked video modeling is a more effective generative paradigm for our dataset and compute constraints. The results consistently show that parameter sharing, particularly in the Full Sharing variant, provides substantial efficiency gains with minimal loss in visual fidelity, very benefitial in resource-constrained settings.

\section{Conclusion}
Humanoid World Models demonstrate that it is possible to build physically plausible, efficient predictive models tailored for humanoid robotics using modest computational resources. Through effective parameter-sharing strategies and compressive VAEs, HWM enables open-world reasoning for embodied agents in compute-constrained settings.

\section*{Impact Statement}

This work advances video-based world models for humanoid robots, aiming to make predictive simulation more computationally accessible. While our models support progress in embodied AI, they are not intended for unsafe or unethical deployment.

\nocite{langley00}

\bibliography{bibliography}

\begin{thebibliography}{61}
\providecommand{\natexlab}[1]{#1}
\providecommand{\url}[1]{\texttt{#1}}
\expandafter\ifx\csname urlstyle\endcsname\relax
  \providecommand{\doi}[1]{doi: #1}\else
  \providecommand{\doi}{doi: \begingroup \urlstyle{rm}\Url}\fi

\bibitem[{1X Technologies}(2024)]{1X_Technologies_1X_World_Model_2024}
{1X Technologies}.
\newblock {1X World Model Challenge}, June 2024.

\bibitem[Albergo et~al.(2023)Albergo, Boffi, and Vanden-Eijnden]{albergo2023stochasticinterpolantsunifyingframework}
Albergo, M.~S., Boffi, N.~M., and Vanden-Eijnden, E.
\newblock Stochastic interpolants: A unifying framework for flows and diffusions, 2023.
\newblock URL \url{https://arxiv.org/abs/2303.08797}.

\bibitem[Bar et~al.(2025)Bar, Zhou, Tran, Darrell, and LeCun]{bar2025navigationworldmodels}
Bar, A., Zhou, G., Tran, D., Darrell, T., and LeCun, Y.
\newblock Navigation world models, 2025.
\newblock URL \url{https://arxiv.org/abs/2412.03572}.

\bibitem[Bar-Tal et~al.(2024)Bar-Tal, Chefer, Tov, Herrmann, Paiss, Zada, Ephrat, Hur, Liu, Raj, Li, Rubinstein, Michaeli, Wang, Sun, Dekel, and Mosseri]{bartal2024lumierespacetimediffusionmodel}
Bar-Tal, O., Chefer, H., Tov, O., Herrmann, C., Paiss, R., Zada, S., Ephrat, A., Hur, J., Liu, G., Raj, A., Li, Y., Rubinstein, M., Michaeli, T., Wang, O., Sun, D., Dekel, T., and Mosseri, I.
\newblock Lumiere: A space-time diffusion model for video generation, 2024.
\newblock URL \url{https://arxiv.org/abs/2401.12945}.

\bibitem[Black et~al.(2024)Black, Brown, Driess, Esmail, Equi, Finn, Fusai, Groom, Hausman, Ichter, Jakubczak, Jones, Ke, Levine, Li-Bell, Mothukuri, Nair, Pertsch, Shi, Tanner, Vuong, Walling, Wang, and Zhilinsky]{black2024pi0}
Black, K., Brown, N., Driess, D., Esmail, A., Equi, M., Finn, C., Fusai, N., Groom, L., Hausman, K., Ichter, B., Jakubczak, S., Jones, T., Ke, L., Levine, S., Li-Bell, A., Mothukuri, M., Nair, S., Pertsch, K., Shi, L.~X., Tanner, J., Vuong, Q., Walling, A., Wang, H., and Zhilinsky, U.
\newblock pi0: A vision-language-action flow model for general robot control.
\newblock \emph{arXiv preprint arXiv:2410.24164}, 2024.
\newblock URL \url{https://arxiv.org/abs/2410.24164}.

\bibitem[Bommasani et~al.(2022)Bommasani, Hudson, Adeli, Altman, Arora, von Arx, Bernstein, Bohg, Bosselut, Brunskill, Brynjolfsson, Buch, Card, Castellon, Chatterji, Chen, Creel, Davis, Demszky, Donahue, Doumbouya, Durmus, Ermon, Etchemendy, Ethayarajh, Fei-Fei, Finn, Gale, Gillespie, Goel, Goodman, Grossman, Guha, Hashimoto, Henderson, Hewitt, Ho, Hong, Hsu, Huang, Icard, Jain, Jurafsky, Kalluri, Karamcheti, Keeling, Khani, Khattab, Koh, Krass, Krishna, Kuditipudi, Kumar, Ladhak, Lee, Lee, Leskovec, Levent, Li, Li, Ma, Malik, Manning, Mirchandani, Mitchell, Munyikwa, Nair, Narayan, Narayanan, Newman, Nie, Niebles, Nilforoshan, Nyarko, Ogut, Orr, Papadimitriou, Park, Piech, Portelance, Potts, Raghunathan, Reich, Ren, Rong, Roohani, Ruiz, Ryan, Ré, Sadigh, Sagawa, Santhanam, Shih, Srinivasan, Tamkin, Taori, Thomas, Tramèr, Wang, Wang, Wu, Wu, Wu, Xie, Yasunaga, You, Zaharia, Zhang, Zhang, Zhang, Zhang, Zheng, Zhou, and Liang]{bommasani2022opportunitiesrisksfoundationmodels}
Bommasani, R., Hudson, D.~A., Adeli, E., Altman, R., Arora, S., von Arx, S., Bernstein, M.~S., Bohg, J., Bosselut, A., Brunskill, E., Brynjolfsson, E., Buch, S., Card, D., Castellon, R., Chatterji, N., Chen, A., Creel, K., Davis, J.~Q., Demszky, D., Donahue, C., Doumbouya, M., Durmus, E., Ermon, S., Etchemendy, J., Ethayarajh, K., Fei-Fei, L., Finn, C., Gale, T., Gillespie, L., Goel, K., Goodman, N., Grossman, S., Guha, N., Hashimoto, T., Henderson, P., Hewitt, J., Ho, D.~E., Hong, J., Hsu, K., Huang, J., Icard, T., Jain, S., Jurafsky, D., Kalluri, P., Karamcheti, S., Keeling, G., Khani, F., Khattab, O., Koh, P.~W., Krass, M., Krishna, R., Kuditipudi, R., Kumar, A., Ladhak, F., Lee, M., Lee, T., Leskovec, J., Levent, I., Li, X.~L., Li, X., Ma, T., Malik, A., Manning, C.~D., Mirchandani, S., Mitchell, E., Munyikwa, Z., Nair, S., Narayan, A., Narayanan, D., Newman, B., Nie, A., Niebles, J.~C., Nilforoshan, H., Nyarko, J., Ogut, G., Orr, L., Papadimitriou, I., Park, J.~S., Piech, C., Portelance, E., Potts, C.,
  Raghunathan, A., Reich, R., Ren, H., Rong, F., Roohani, Y., Ruiz, C., Ryan, J., Ré, C., Sadigh, D., Sagawa, S., Santhanam, K., Shih, A., Srinivasan, K., Tamkin, A., Taori, R., Thomas, A.~W., Tramèr, F., Wang, R.~E., Wang, W., Wu, B., Wu, J., Wu, Y., Xie, S.~M., Yasunaga, M., You, J., Zaharia, M., Zhang, M., Zhang, T., Zhang, X., Zhang, Y., Zheng, L., Zhou, K., and Liang, P.
\newblock On the opportunities and risks of foundation models, 2022.
\newblock URL \url{https://arxiv.org/abs/2108.07258}.

\bibitem[Bruce et~al.(2024)Bruce, Dennis, Edwards, Parker-Holder, Shi, Hughes, Lai, Mavalankar, Steigerwald, Apps, et~al.]{bruce2024genie}
Bruce, J., Dennis, M.~D., Edwards, A., Parker-Holder, J., Shi, Y., Hughes, E., Lai, M., Mavalankar, A., Steigerwald, R., Apps, C., et~al.
\newblock Genie: Generative interactive environments.
\newblock In \emph{Forty-first International Conference on Machine Learning}, 2024.

\bibitem[Chang et~al.(2022{\natexlab{a}})Chang, Zhang, Jiang, Liu, and Freeman]{chang2022maskgit}
Chang, H., Zhang, H., Jiang, L., Liu, C., and Freeman, W.~T.
\newblock Maskgit: Masked generative image transformer.
\newblock In \emph{Proceedings of the IEEE/CVF conference on computer vision and pattern recognition}, pp.\  11315--11325, 2022{\natexlab{a}}.

\bibitem[Chang et~al.(2022{\natexlab{b}})Chang, Zhang, Jiang, Liu, and Freeman]{chang2022maskgitmaskedgenerativeimage}
Chang, H., Zhang, H., Jiang, L., Liu, C., and Freeman, W.~T.
\newblock Maskgit: Masked generative image transformer, 2022{\natexlab{b}}.
\newblock URL \url{https://arxiv.org/abs/2202.04200}.

\bibitem[Chen et~al.(2025)Chen, Qian, Hu, Fu, Tong, Wang, Li, Zhang, Schwing, Liu, and Yang]{chen2025ditairrevisitingefficiencydiffusion}
Chen, C., Qian, R., Hu, W., Fu, T.-J., Tong, J., Wang, X., Li, L., Zhang, B., Schwing, A., Liu, W., and Yang, Y.
\newblock Dit-air: Revisiting the efficiency of diffusion model architecture design in text to image generation, 2025.
\newblock URL \url{https://arxiv.org/abs/2503.10618}.

\bibitem[Du et~al.(2023)Du, Yang, Florence, Xia, Wahid, Ichter, Sermanet, Yu, Abbeel, Tenenbaum, Kaelbling, Zeng, and Tompson]{du2023videolanguageplanning}
Du, Y., Yang, M., Florence, P., Xia, F., Wahid, A., Ichter, B., Sermanet, P., Yu, T., Abbeel, P., Tenenbaum, J.~B., Kaelbling, L., Zeng, A., and Tompson, J.
\newblock Video language planning, 2023.
\newblock URL \url{https://arxiv.org/abs/2310.10625}.

\bibitem[Duan et~al.(2024)Duan, Pumacay, Kumar, Wang, Tian, Yuan, Krishna, Fox, Mandlekar, and Guo]{duan2024ahavisionlanguagemodeldetectingreasoning}
Duan, J., Pumacay, W., Kumar, N., Wang, Y.~R., Tian, S., Yuan, W., Krishna, R., Fox, D., Mandlekar, A., and Guo, Y.
\newblock Aha: A vision-language-model for detecting and reasoning over failures in robotic manipulation, 2024.
\newblock URL \url{https://arxiv.org/abs/2410.00371}.

\bibitem[Esser et~al.(2024)Esser, Kulal, Blattmann, Entezari, Müller, Saini, Levi, Lorenz, Sauer, Boesel, Podell, Dockhorn, English, Lacey, Goodwin, Marek, and Rombach]{esser2024scalingrectifiedflowtransformers}
Esser, P., Kulal, S., Blattmann, A., Entezari, R., Müller, J., Saini, H., Levi, Y., Lorenz, D., Sauer, A., Boesel, F., Podell, D., Dockhorn, T., English, Z., Lacey, K., Goodwin, A., Marek, Y., and Rombach, R.
\newblock Scaling rectified flow transformers for high-resolution image synthesis, 2024.
\newblock URL \url{https://arxiv.org/abs/2403.03206}.

\bibitem[fal.ai Blog(2024)]{fal_auraflow}
fal.ai Blog.
\newblock Auraflow: Generate high-fidelity 3d assets with diffusion models.
\newblock \url{https://blog.fal.ai/auraflow/}, Apr 2024.
\newblock Accessed April 19, 2025.

\bibitem[Goodfellow et~al.(2014)Goodfellow, Pouget-Abadie, Mirza, Xu, Warde-Farley, Ozair, Courville, and Bengio]{goodfellow2014generativeadversarialnetworks}
Goodfellow, I.~J., Pouget-Abadie, J., Mirza, M., Xu, B., Warde-Farley, D., Ozair, S., Courville, A., and Bengio, Y.
\newblock Generative adversarial networks, 2014.
\newblock URL \url{https://arxiv.org/abs/1406.2661}.

\bibitem[Goswami \& Vadakkepat(2019)Goswami and Vadakkepat]{humanoid_ref}
Goswami, A. and Vadakkepat, P. (eds.).
\newblock \emph{Humanoid Robotics: A Reference}.
\newblock Springer Dordrecht, 2019.
\newblock ISBN 978-94-007-6046-2.
\newblock \doi{10.1007/978-94-007-6046-2}.
\newblock URL \url{https://link.springer.com/referencework/10.1007/978-94-007-6046-2}.

\bibitem[Ha \& Schmidhuber(2018)Ha and Schmidhuber]{haworldmodel}
Ha, D. and Schmidhuber, J.
\newblock World models.
\newblock \emph{CoRR}, abs/1803.10122, 2018.
\newblock URL \url{http://dblp.uni-trier.de/db/journals/corr/corr1803.html#abs-1803-10122}.

\bibitem[Harvey et~al.(2022)Harvey, Naderiparizi, and Wood]{harvey2022conditionalimagegenerationconditioning}
Harvey, W., Naderiparizi, S., and Wood, F.
\newblock Conditional image generation by conditioning variational auto-encoders, 2022.
\newblock URL \url{https://arxiv.org/abs/2102.12037}.

\bibitem[Heusel et~al.(2018)Heusel, Ramsauer, Unterthiner, Nessler, and Hochreiter]{fid}
Heusel, M., Ramsauer, H., Unterthiner, T., Nessler, B., and Hochreiter, S.
\newblock Gans trained by a two time-scale update rule converge to a local nash equilibrium, 2018.
\newblock URL \url{https://arxiv.org/abs/1706.08500}.

\bibitem[Hirose \& Ogawa(2007)Hirose and Ogawa]{humanoid_industrial}
Hirose, M. and Ogawa, K.
\newblock Honda humanoid robots development.
\newblock \emph{Philosophical Transactions of the Royal Society A: Mathematical, Physical and Engineering Sciences}, 365\penalty0 (1850):\penalty0 11--19, 2007.
\newblock \doi{10.1098/rsta.2006.1917}.
\newblock URL \url{https://royalsocietypublishing.org/doi/10.1098/rsta.2006.1917}.

\bibitem[Ho \& Salimans(2022)Ho and Salimans]{ho2022classifierfreediffusionguidance}
Ho, J. and Salimans, T.
\newblock Classifier-free diffusion guidance, 2022.
\newblock URL \url{https://arxiv.org/abs/2207.12598}.

\bibitem[Ho et~al.(2020)Ho, Jain, and Abbeel]{ho2020denoisingdiffusionprobabilisticmodels}
Ho, J., Jain, A., and Abbeel, P.
\newblock Denoising diffusion probabilistic models, 2020.
\newblock URL \url{https://arxiv.org/abs/2006.11239}.

\bibitem[Ho et~al.(2022)Ho, Salimans, Gritsenko, Chan, Norouzi, and Fleet]{ho2022videodiffusionmodels}
Ho, J., Salimans, T., Gritsenko, A., Chan, W., Norouzi, M., and Fleet, D.~J.
\newblock Video diffusion models, 2022.
\newblock URL \url{https://arxiv.org/abs/2204.03458}.

\bibitem[Horé \& Ziou(2010)Horé and Ziou]{psnr}
Horé, A. and Ziou, D.
\newblock Image quality metrics: Psnr vs. ssim.
\newblock In \emph{2010 20th International Conference on Pattern Recognition}, pp.\  2366--2369, 2010.
\newblock \doi{10.1109/ICPR.2010.579}.

\bibitem[Imtiaz \& Khan(2024)Imtiaz and Khan]{humanoid_in_home}
Imtiaz, R. and Khan, A.
\newblock Perceptions of humanoid robots in caregiving: A study of skilled nursing home and long term care administrators, 2024.
\newblock URL \url{https://arxiv.org/abs/2401.02105}.

\bibitem[Jin et~al.(2024)Jin, Sun, Li, Xu, Xu, Jiang, Zhuang, Huang, Song, Mu, and Lin]{jin2024pyramidalflowmatchingefficient}
Jin, Y., Sun, Z., Li, N., Xu, K., Xu, K., Jiang, H., Zhuang, N., Huang, Q., Song, Y., Mu, Y., and Lin, Z.
\newblock Pyramidal flow matching for efficient video generative modeling, 2024.
\newblock URL \url{https://arxiv.org/abs/2410.05954}.

\bibitem[Kim et~al.(2024)Kim, Pertsch, Karamcheti, Xiao, Balakrishna, Nair, Rafailov, Foster, Lam, Sanketi, Vuong, Kollar, Burchfiel, Tedrake, Sadigh, Levine, Liang, and Finn]{kim24openvla}
Kim, M., Pertsch, K., Karamcheti, S., Xiao, T., Balakrishna, A., Nair, S., Rafailov, R., Foster, E., Lam, G., Sanketi, P., Vuong, Q., Kollar, T., Burchfiel, B., Tedrake, R., Sadigh, D., Levine, S., Liang, P., and Finn, C.
\newblock Openvla: An open-source vision-language-action model.
\newblock \emph{arXiv preprint arXiv:2406.09246}, 2024.

\bibitem[Kong et~al.(2025)Kong, Tian, Zhang, Min, Dai, Zhou, Xiong, Li, Wu, Zhang, Wu, Lin, Yuan, Long, Wang, Wang, Li, Huang, Yang, Tan, Wang, Song, Bai, Wu, Xue, Wang, Wang, Liu, Li, Li, Wang, Yu, Deng, Li, Chen, Cui, Peng, Yu, He, Xu, Zhou, Xu, Tao, Lu, Liu, Zhou, Wang, Yang, Wang, Liu, Jiang, and Zhong]{kong2025hunyuanvideosystematicframeworklarge}
Kong, W., Tian, Q., Zhang, Z., Min, R., Dai, Z., Zhou, J., Xiong, J., Li, X., Wu, B., Zhang, J., Wu, K., Lin, Q., Yuan, J., Long, Y., Wang, A., Wang, A., Li, C., Huang, D., Yang, F., Tan, H., Wang, H., Song, J., Bai, J., Wu, J., Xue, J., Wang, J., Wang, K., Liu, M., Li, P., Li, S., Wang, W., Yu, W., Deng, X., Li, Y., Chen, Y., Cui, Y., Peng, Y., Yu, Z., He, Z., Xu, Z., Zhou, Z., Xu, Z., Tao, Y., Lu, Q., Liu, S., Zhou, D., Wang, H., Yang, Y., Wang, D., Liu, Y., Jiang, J., and Zhong, C.
\newblock Hunyuanvideo: A systematic framework for large video generative models, 2025.
\newblock URL \url{https://arxiv.org/abs/2412.03603}.

\bibitem[Langley(2000)]{langley00}
Langley, P.
\newblock Crafting papers on machine learning.
\newblock In Langley, P. (ed.), \emph{Proceedings of the 17th International Conference on Machine Learning (ICML 2000)}, pp.\  1207--1216, Stanford, CA, 2000. Morgan Kaufmann.

\bibitem[Li et~al.(2024)Li, Yu, She, Yu, Lan, Zhu, Hu, and Xu]{li2024llmenhancedscenegraphlearning}
Li, W., Yu, Z., She, Q., Yu, Z., Lan, Y., Zhu, C., Hu, R., and Xu, K.
\newblock Llm-enhanced scene graph learning for household rearrangement, 2024.
\newblock URL \url{https://arxiv.org/abs/2408.12093}.

\bibitem[Liang et~al.(2023)Liang, Huang, Xia, Xu, Hausman, Ichter, Florence, and Zeng]{liang2023codepolicieslanguagemodel}
Liang, J., Huang, W., Xia, F., Xu, P., Hausman, K., Ichter, B., Florence, P., and Zeng, A.
\newblock Code as policies: Language model programs for embodied control, 2023.
\newblock URL \url{https://arxiv.org/abs/2209.07753}.

\bibitem[Lipman et~al.(2023)Lipman, Chen, Ben-Hamu, Nickel, and Le]{lipman2023flowmatchinggenerativemodeling}
Lipman, Y., Chen, R. T.~Q., Ben-Hamu, H., Nickel, M., and Le, M.
\newblock Flow matching for generative modeling, 2023.
\newblock URL \url{https://arxiv.org/abs/2210.02747}.

\bibitem[Liu et~al.(2022)Liu, Gong, and Liu]{liu2022flowstraightfastlearning}
Liu, X., Gong, C., and Liu, Q.
\newblock Flow straight and fast: Learning to generate and transfer data with rectified flow, 2022.
\newblock URL \url{https://arxiv.org/abs/2209.03003}.

\bibitem[Liu et~al.(2024)Liu, Zhang, Li, Yan, Gao, Chen, Yuan, Huang, Sun, Gao, He, and Sun]{liu2024sorareviewbackgroundtechnology}
Liu, Y., Zhang, K., Li, Y., Yan, Z., Gao, C., Chen, R., Yuan, Z., Huang, Y., Sun, H., Gao, J., He, L., and Sun, L.
\newblock Sora: A review on background, technology, limitations, and opportunities of large vision models, 2024.
\newblock URL \url{https://arxiv.org/abs/2402.17177}.

\bibitem[Luo et~al.(2024)Luo, Shi, Ge, Yang, Wang, and Shan]{luo2024openmagvit2opensourceprojectdemocratizing}
Luo, Z., Shi, F., Ge, Y., Yang, Y., Wang, L., and Shan, Y.
\newblock Open-magvit2: An open-source project toward democratizing auto-regressive visual generation, 2024.
\newblock URL \url{https://arxiv.org/abs/2409.04410}.

\bibitem[NVIDIA et~al.(2025)NVIDIA, :, Agarwal, Ali, Bala, Balaji, Barker, Cai, Chattopadhyay, Chen, Cui, Ding, Dworakowski, Fan, Fenzi, Ferroni, Fidler, Fox, Ge, Ge, Gu, Gururani, He, Huang, Huffman, Jannaty, Jin, Kim, Klár, Lam, Lan, Leal-Taixe, Li, Li, Lin, Lin, Ling, Liu, Liu, Luo, Ma, Mao, Mo, Mousavian, Nah, Niverty, Page, Paschalidou, Patel, Pavao, Ramezanali, Reda, Ren, Sabavat, Schmerling, Shi, Stefaniak, Tang, Tchapmi, Tredak, Tseng, Varghese, Wang, Wang, Wang, Wang, Wei, Wei, Wu, Xu, Yang, Yen-Chen, Zeng, Zeng, Zhang, Zhang, Zhang, Zhao, and Zolkowski]{nvidia2025cosmosworldfoundationmodel}
NVIDIA, :, Agarwal, N., Ali, A., Bala, M., Balaji, Y., Barker, E., Cai, T., Chattopadhyay, P., Chen, Y., Cui, Y., Ding, Y., Dworakowski, D., Fan, J., Fenzi, M., Ferroni, F., Fidler, S., Fox, D., Ge, S., Ge, Y., Gu, J., Gururani, S., He, E., Huang, J., Huffman, J., Jannaty, P., Jin, J., Kim, S.~W., Klár, G., Lam, G., Lan, S., Leal-Taixe, L., Li, A., Li, Z., Lin, C.-H., Lin, T.-Y., Ling, H., Liu, M.-Y., Liu, X., Luo, A., Ma, Q., Mao, H., Mo, K., Mousavian, A., Nah, S., Niverty, S., Page, D., Paschalidou, D., Patel, Z., Pavao, L., Ramezanali, M., Reda, F., Ren, X., Sabavat, V. R.~N., Schmerling, E., Shi, S., Stefaniak, B., Tang, S., Tchapmi, L., Tredak, P., Tseng, W.-C., Varghese, J., Wang, H., Wang, H., Wang, H., Wang, T.-C., Wei, F., Wei, X., Wu, J.~Z., Xu, J., Yang, W., Yen-Chen, L., Zeng, X., Zeng, Y., Zhang, J., Zhang, Q., Zhang, Y., Zhao, Q., and Zolkowski, A.
\newblock Cosmos world foundation model platform for physical ai, 2025.
\newblock URL \url{https://arxiv.org/abs/2501.03575}.

\bibitem[Peebles \& Xie(2023)Peebles and Xie]{peebles2023scalablediffusionmodelstransformers}
Peebles, W. and Xie, S.
\newblock Scalable diffusion models with transformers, 2023.
\newblock URL \url{https://arxiv.org/abs/2212.09748}.

\bibitem[Pfeifer \& Iida(2004)Pfeifer and Iida]{embodiedai_challenges}
Pfeifer, R. and Iida, F.
\newblock \emph{Embodied Artificial Intelligence: Trends and Challenges}, pp.\  1--26.
\newblock Springer Berlin Heidelberg, Berlin, Heidelberg, 2004.
\newblock ISBN 978-3-540-27833-7.
\newblock \doi{10.1007/978-3-540-27833-7_1}.
\newblock URL \url{https://doi.org/10.1007/978-3-540-27833-7_1}.

\bibitem[Polyak et~al.(2024)Polyak, Zohar, Brown, Tjandra, Sinha, Lee, Vyas, Shi, Ma, Chuang, Yan, Choudhary, Wang, Sethi, Pang, Ma, Misra, Hou, Wang, Jagadeesh, Li, Zhang, Singh, Williamson, Le, Yu, Singh, Zhang, Vajda, Duval, Girdhar, Sumbaly, Rambhatla, Tsai, Azadi, Datta, Chen, Bell, Ramaswamy, Sheynin, Bhattacharya, Motwani, Xu, Li, Hou, Hsu, Yin, Dai, Taigman, Luo, Liu, Wu, Zhao, Kirstain, He, He, Pumarola, Thabet, Sanakoyeu, Mallya, Guo, Araya, Kerr, Wood, Liu, Peng, Vengertsev, Schonfeld, Blanchard, Juefei-Xu, Nord, Liang, Hoffman, Kohler, Fire, Sivakumar, Chen, Yu, Gao, Georgopoulos, Moritz, Sampson, Li, Parmeggiani, Fine, Fowler, Petrovic, and Du]{polyak2024moviegencastmedia}
Polyak, A., Zohar, A., Brown, A., Tjandra, A., Sinha, A., Lee, A., Vyas, A., Shi, B., Ma, C.-Y., Chuang, C.-Y., Yan, D., Choudhary, D., Wang, D., Sethi, G., Pang, G., Ma, H., Misra, I., Hou, J., Wang, J., Jagadeesh, K., Li, K., Zhang, L., Singh, M., Williamson, M., Le, M., Yu, M., Singh, M.~K., Zhang, P., Vajda, P., Duval, Q., Girdhar, R., Sumbaly, R., Rambhatla, S.~S., Tsai, S., Azadi, S., Datta, S., Chen, S., Bell, S., Ramaswamy, S., Sheynin, S., Bhattacharya, S., Motwani, S., Xu, T., Li, T., Hou, T., Hsu, W.-N., Yin, X., Dai, X., Taigman, Y., Luo, Y., Liu, Y.-C., Wu, Y.-C., Zhao, Y., Kirstain, Y., He, Z., He, Z., Pumarola, A., Thabet, A., Sanakoyeu, A., Mallya, A., Guo, B., Araya, B., Kerr, B., Wood, C., Liu, C., Peng, C., Vengertsev, D., Schonfeld, E., Blanchard, E., Juefei-Xu, F., Nord, F., Liang, J., Hoffman, J., Kohler, J., Fire, K., Sivakumar, K., Chen, L., Yu, L., Gao, L., Georgopoulos, M., Moritz, R., Sampson, S.~K., Li, S., Parmeggiani, S., Fine, S., Fowler, T., Petrovic, V., and Du, Y.
\newblock Movie gen: A cast of media foundation models, 2024.
\newblock URL \url{https://arxiv.org/abs/2410.13720}.

\bibitem[Ramesh et~al.(2021)Ramesh, Pavlov, Goh, Gray, Voss, Radford, Chen, and Sutskever]{ramesh2021zeroshottexttoimagegeneration}
Ramesh, A., Pavlov, M., Goh, G., Gray, S., Voss, C., Radford, A., Chen, M., and Sutskever, I.
\newblock Zero-shot text-to-image generation, 2021.
\newblock URL \url{https://arxiv.org/abs/2102.12092}.

\bibitem[Rombach et~al.(2022)Rombach, Blattmann, Lorenz, Esser, and Ommer]{rombach2022highresolutionimagesynthesislatent}
Rombach, R., Blattmann, A., Lorenz, D., Esser, P., and Ommer, B.
\newblock High-resolution image synthesis with latent diffusion models, 2022.
\newblock URL \url{https://arxiv.org/abs/2112.10752}.

\bibitem[Sohl-Dickstein et~al.(2015)Sohl-Dickstein, Weiss, Maheswaranathan, and Ganguli]{sohldickstein2015deepunsupervisedlearningusing}
Sohl-Dickstein, J., Weiss, E.~A., Maheswaranathan, N., and Ganguli, S.
\newblock Deep unsupervised learning using nonequilibrium thermodynamics, 2015.
\newblock URL \url{https://arxiv.org/abs/1503.03585}.

\bibitem[Song et~al.(2021)Song, Sohl-Dickstein, Kingma, Kumar, Ermon, and Poole]{song2021scorebasedgenerativemodelingstochastic}
Song, Y., Sohl-Dickstein, J., Kingma, D.~P., Kumar, A., Ermon, S., and Poole, B.
\newblock Score-based generative modeling through stochastic differential equations, 2021.
\newblock URL \url{https://arxiv.org/abs/2011.13456}.

\bibitem[Su et~al.(2023)Su, Lu, Pan, Murtadha, Wen, and Liu]{su2023roformerenhancedtransformerrotary}
Su, J., Lu, Y., Pan, S., Murtadha, A., Wen, B., and Liu, Y.
\newblock Roformer: Enhanced transformer with rotary position embedding, 2023.
\newblock URL \url{https://arxiv.org/abs/2104.09864}.

\bibitem[Tong et~al.(2024)Tong, Liu, Zhai, Ma, LeCun, and Xie]{tong2024eyeswideshutexploring}
Tong, S., Liu, Z., Zhai, Y., Ma, Y., LeCun, Y., and Xie, S.
\newblock Eyes wide shut? exploring the visual shortcomings of multimodal llms, 2024.
\newblock URL \url{https://arxiv.org/abs/2401.06209}.

\bibitem[van~den Oord et~al.(2018)van~den Oord, Vinyals, and Kavukcuoglu]{oord2018neuraldiscreterepresentationlearning}
van~den Oord, A., Vinyals, O., and Kavukcuoglu, K.
\newblock Neural discrete representation learning, 2018.
\newblock URL \url{https://arxiv.org/abs/1711.00937}.

\bibitem[Vaswani et~al.(2023)Vaswani, Shazeer, Parmar, Uszkoreit, Jones, Gomez, Kaiser, and Polosukhin]{vaswani2023attentionneed}
Vaswani, A., Shazeer, N., Parmar, N., Uszkoreit, J., Jones, L., Gomez, A.~N., Kaiser, L., and Polosukhin, I.
\newblock Attention is all you need, 2023.
\newblock URL \url{https://arxiv.org/abs/1706.03762}.

\bibitem[Vianello et~al.(2021)Vianello, Penco, Gomes, You, Anzalone, Maurice, Thomas, and Ivaldi]{humanoid_interaction}
Vianello, L., Penco, L., Gomes, W., You, Y., Anzalone, S.~M., Maurice, P., Thomas, V., and Ivaldi, S.
\newblock Human-humanoid interaction and cooperation: a review.
\newblock \emph{Current Robotics Reports}, 2\penalty0 (4):\penalty0 441--454, December 2021.
\newblock \doi{10.1007/s43154-021-00068-z}.
\newblock URL \url{https://doi.org/10.1007/s43154-021-00068-z}.

\bibitem[Wang et~al.(2024)Wang, Han, Jiao, Zhang, Wu, Zhu, and Liu]{wang2024llm3largelanguagemodelbasedtask}
Wang, S., Han, M., Jiao, Z., Zhang, Z., Wu, Y.~N., Zhu, S.-C., and Liu, H.
\newblock Llm3:large language model-based task and motion planning with motion failure reasoning, 2024.
\newblock URL \url{https://arxiv.org/abs/2403.11552}.

\bibitem[Wu et~al.(2024)Wu, Yin, Feng, He, Li, Hao, and Long]{wu2024ivideogpt}
Wu, J., Yin, S., Feng, N., He, X., Li, D., Hao, J., and Long, M.
\newblock ivideogpt: Interactive videogpts are scalable world models.
\newblock In \emph{Advances in Neural Information Processing Systems}, 2024.

\bibitem[Xiang et~al.(2024)Xiang, Liu, Gu, Gao, Ning, Zha, Feng, Tao, Hao, Shi, Liu, Xing, and Hu]{xiang2024pandorageneralworldmodel}
Xiang, J., Liu, G., Gu, Y., Gao, Q., Ning, Y., Zha, Y., Feng, Z., Tao, T., Hao, S., Shi, Y., Liu, Z., Xing, E.~P., and Hu, Z.
\newblock Pandora: Towards general world model with natural language actions and video states, 2024.
\newblock URL \url{https://arxiv.org/abs/2406.09455}.

\bibitem[Xing et~al.(2024)Xing, Feng, Chen, Dai, Hu, Xu, Wu, and Jiang]{surveyvideodiffusionmodels}
Xing, Z., Feng, Q., Chen, H., Dai, Q., Hu, H., Xu, H., Wu, Z., and Jiang, Y.-G.
\newblock A survey on video diffusion models, 2024.
\newblock URL \url{https://arxiv.org/abs/2310.10647}.

\bibitem[Yang et~al.(2023)Yang, Du, Ghasemipour, Tompson, Schuurmans, and Abbeel]{unisim}
Yang, M., Du, Y., Ghasemipour, K., Tompson, J., Schuurmans, D., and Abbeel, P.
\newblock Learning interactive real-world simulators.
\newblock \emph{arXiv preprint arXiv:2310.06114}, 2023.

\bibitem[Yang et~al.(2024)Yang, Walker, Parker-Holder, Du, Bruce, Barreto, Abbeel, and Schuurmans]{yang2024videonewlanguagerealworld}
Yang, S., Walker, J., Parker-Holder, J., Du, Y., Bruce, J., Barreto, A., Abbeel, P., and Schuurmans, D.
\newblock Video as the new language for real-world decision making, 2024.
\newblock URL \url{https://arxiv.org/abs/2402.17139}.

\bibitem[Yang et~al.(2025)Yang, Teng, Zheng, Ding, Huang, Xu, Yang, Hong, Zhang, Feng, Yin, Zhang, Wang, Cheng, Xu, Gu, Dong, and Tang]{yang2025cogvideoxtexttovideodiffusionmodels}
Yang, Z., Teng, J., Zheng, W., Ding, M., Huang, S., Xu, J., Yang, Y., Hong, W., Zhang, X., Feng, G., Yin, D., Zhang, Y., Wang, W., Cheng, Y., Xu, B., Gu, X., Dong, Y., and Tang, J.
\newblock Cogvideox: Text-to-video diffusion models with an expert transformer, 2025.
\newblock URL \url{https://arxiv.org/abs/2408.06072}.

\bibitem[Yu et~al.(2023)Yu, Cheng, Sohn, Lezama, Zhang, Chang, Hauptmann, Yang, Hao, Essa, and Jiang]{yu2023magvitmaskedgenerativevideo}
Yu, L., Cheng, Y., Sohn, K., Lezama, J., Zhang, H., Chang, H., Hauptmann, A.~G., Yang, M.-H., Hao, Y., Essa, I., and Jiang, L.
\newblock Magvit: Masked generative video transformer, 2023.
\newblock URL \url{https://arxiv.org/abs/2212.05199}.

\bibitem[Yu et~al.(2024)Yu, Lezama, Gundavarapu, Versari, Sohn, Minnen, Cheng, Birodkar, Gupta, Gu, Hauptmann, Gong, Yang, Essa, Ross, and Jiang]{yu2024languagemodelbeatsdiffusion}
Yu, L., Lezama, J., Gundavarapu, N.~B., Versari, L., Sohn, K., Minnen, D., Cheng, Y., Birodkar, V., Gupta, A., Gu, X., Hauptmann, A.~G., Gong, B., Yang, M.-H., Essa, I., Ross, D.~A., and Jiang, L.
\newblock Language model beats diffusion -- tokenizer is key to visual generation, 2024.
\newblock URL \url{https://arxiv.org/abs/2310.05737}.

\bibitem[Zhang et~al.(2024)Zhang, Huang, Jin, and Lu]{zhang2024visionlanguagemodelsvisiontasks}
Zhang, J., Huang, J., Jin, S., and Lu, S.
\newblock Vision-language models for vision tasks: A survey, 2024.
\newblock URL \url{https://arxiv.org/abs/2304.00685}.

\bibitem[Zhang et~al.(2023)Zhang, Xiong, Yang, Casas, Hu, and Urtasun]{zhang2023copilot4d}
Zhang, L., Xiong, Y., Yang, Z., Casas, S., Hu, R., and Urtasun, R.
\newblock Copilot4d: Learning unsupervised world models for autonomous driving via discrete diffusion.
\newblock \emph{arXiv preprint arXiv:2311.01017}, 2023.

\bibitem[Zhao et~al.(2023)Zhao, Kumar, Levine, and Finn]{zhao2023learningfinegrainedbimanualmanipulation}
Zhao, T.~Z., Kumar, V., Levine, S., and Finn, C.
\newblock Learning fine-grained bimanual manipulation with low-cost hardware, 2023.
\newblock URL \url{https://arxiv.org/abs/2304.13705}.

\bibitem[Zhu et~al.(2024)Zhu, Wu, Guo, Liu, Cheang, and Kong]{FangqiIRASim2024}
Zhu, F., Wu, H., Guo, S., Liu, Y., Cheang, C., and Kong, T.
\newblock Irasim: Learning interactive real-robot action simulators.
\newblock \emph{arXiv:2406.12802}, 2024.

\end{thebibliography}
\bibliographystyle{icml2025}

\newpage
\appendix
\onecolumn



\end{document}